\documentclass[lettersize,journal]{IEEEtran}
\usepackage{amsmath,amsfonts}
\usepackage{algorithmic}
\usepackage{algorithm}
\usepackage{array}
\usepackage[caption=false,font=normalsize,labelfont=sf,textfont=sf]{subfig}
\usepackage{textcomp}
\usepackage{stfloats}
\usepackage{url}
\usepackage{verbatim}
\usepackage{graphicx}
\usepackage{cite}
\begin{document}

\title{On the importance of stationarity, strong baselines and benchmarks in transport prediction problems}

\author{Filipe Rodrigues
\thanks{F.~Rodrigues is with the Technical University of Denmark (DTU), Bygning 115, 2800 Kgs. Lyngby, Denmark. E-mail: rodr@dtu.dk}
}

\markboth{Journal of \LaTeX\ Class Files,~Vol.~14, No.~8, March~2022}%
{Shell \MakeLowercase{\textit{et al.}}: A Sample Article Using IEEEtran.cls for IEEE Journals}


\maketitle

\begin{abstract}
Over the last years, the transportation community has witnessed a tremendous amount of research contributions on new deep learning approaches for spatio-temporal forecasting. These contributions tend to emphasize the modeling of spatial correlations, while neglecting the fairly stable and recurrent nature of human mobility patterns. In this short paper, we show that a naive baseline method based on the average weekly pattern and linear regression can achieve comparable results to many state-of-the-art deep learning approaches for spatio-temporal forecasting in transportation, or even outperform them on several datasets, thus contrasting the importance of stationarity and recurrent patterns in the data with the importance of spatial correlations. Furthermore, we establish 9 different reference benchmarks that can be used to compare new approaches for spatio-temporal forecasting, and provide a discussion on best practices and the direction that the field is taking. 
\end{abstract}


\section{Introduction}

The emerging technologies and new sensors in transportation provide unique opportunities for machine learning to contribute with prediction models that can be used to manage and optimize the transportation system, while also providing users with services that improve their overall experience. This includes data from inductive loop detectors, probe vehicles, smart-card technologies, ride-sharing apps, etc. Since this data is known to exhibit strong correlations in both time and space, a very significant amount of research effort is currently placed on building prediction models that can account for those correlations. Due to the popularity of deep learning methods, the transportation field is currently witnessing a tremendous amount of research contributions on new methods for spatio-temporal prediction problems, which are typically based on deep learning. This includes prediction models for traffic speeds \cite{yu2017spatio,lee2022ddp,zhao2019t,li2018diffusion,yang2021real}, traffic volumes \cite{choi2021graph}, taxi demand \cite{yao2018deep}, bike-sharing demand \cite{ye2020coupled,xia20213dgcn}, railway delays \cite{heglund2020railway}, etc. Recent comprehensive surveys are provided in \cite{ye2020build}, \cite{xie2020urban} and \cite{jiang2021graph}.

While this line of research has brought exciting new ideas on how to capture complex spatial correlation patterns in deep learning models (see e.g., \cite{li2018diffusion,choi2021graph}), the research community working on these problems seems to have forgotten other well-known properties of this type of data. Concretely, it is well known that human mobility patterns are fairly stable and recurrent. Road traffic in a given area on Monday at 8am tends to look very similar every week. Likewise, the general transportation demand pattern in an area tends to repeat itself on a weekly basis (e.g., high demand in the mornings and low demand in the evenings for residential areas on workdays). Nevertheless, a wide proportion of the recent literature on spatio-temporal prediction for transportation problems ignores this. Instead, the forecasting methods considered often operate directly on the highly non-stationary time series data. This contradicts basic knowledge from time series analysis, where the importance of making time series data stationary is strongly emphasized \cite{hyndman2018forecasting,hamilton2020time}. Stationarity is often achieved by transforming the observations (e.g., with logarithm) or through differencing, where the observation at time $t-1$ is subtracted to the observation at time $t$. Other variants exist, such as second-order differencing, or seasonal differencing, where the difference between an observation and the previous observation from the same season is considered \cite{hyndman2018forecasting}. 

Since mobility data typically exhibits a weekly recurring pattern, we argue that a simple and effective approach to make the time series data less non-stationary is to compute the average weekly pattern, remove it from the observations, and model the obtained residuals instead. With this in mind, we downloaded 9 datasets and the respective source code provided by 8 recent publications in top-tier conferences and journals for different transport prediction problems (traffic speeds, traffic volumes, demand prediction, and in- and out-flows), and reproduced their experiments. Then, using the exact same experimental setup, we compared the performance of the proposed approaches to a simple linear regression on the $h$ previous observations that operates on the residuals from the average weekly pattern. Our empirical results show that this naive baseline turns out to be very competitive with extremely complex and computationally-demanding deep-learning-based approaches proposed in the recent literature, and can even outperform several of them in many datasets. These results therefore highlight the need for strong baselines in a field that is currently very prolific in producing incremental research contributions on the spatio-temporal modeling of transportation data using deep learning. This problem is further aggravated by the lack of standard benchmark datasets that can used as a common reference for direct comparisons between approaches. Ultimately, the goal of this short paper is to ask the community to take a step back, question the direction that it is currently taking, and provide recommendations. 

In summary, our contributions are the following:
\begin{itemize}
\item We empirically demonstrate the importance of considering the weekly recurring pattern in the data and how a large fraction of the recent literature yet ignores it;
\item We propose a very simple baseline method for spatio-temporal prediction for transportation problems that takes seconds to compute, requires no hyper-parameter tuning, and can often achieve comparable results to complex state-of-the-art approaches;
\item We establish 9 different benchmark datasets and respective experimental setups that can be used to compare new approaches for spatio-temporal prediction;
\item We provide a discussion on trends and current practices in the field in relation to the ``best practices'' that are adopted in other research communities such as computer vision and natural language processing. 
\end{itemize} 

\section{Methods}
\label{sec:methods}

\begin{table*}[htp]
\caption{Datasets for benchmarking.}
\begin{center}
\begin{tabular}{c|c|c|c|c|c}
Name & Type & Timespan & Time granularity & Train/val/test split & Source\\
\hline
PeMSD7(M) - California & traffic speeds & 01/04/2016 - 30/06/2016 & 5 minutes & 34/5/5 days & Yu et al., 2018 \cite{yu2017spatio} \\
Urban1 - South Korea & traffic speeds & 01/04/2018 - 30/04/2018 & 5 minutes & 70/10/20 \% & Lee and Rhee, 2022 \cite{lee2022ddp} \\
NYC Citi Bike - New York & pickups and dropoffs & 01/04/2016 - 01/04/2016 & 30 minutes & 63/14/14 days & Ye et al., 2021 \cite{ye2020coupled} \\
PeMSD4 - California & traffic volumes & 01/01/2018 - 28/02/2018 & 5 minutes & 60/20/20 \% & Choi et al., 2022 \cite{choi2021graph} \\
SZ-taxi - Shenzhen & traffic speeds & 01/01/2015 - 31/01/2015 & 15 minutes & 80/-/20 \% & Zhao et al., 2021 \cite{zhao2019t} \\
METR-LA - Los Angeles & traffic speeds & 01/03/2012 - 30/06/2012 & 5 minutes & 70/10/20 \% & Li et al., 2018 \cite{li2018diffusion} \\
PEMS-BAY - California & traffic speeds & 01/01/2017 - 31/05/2017 & 5 minutes & 70/10/20 \% & Li et al., 2018 \cite{li2018diffusion} \\
NYC Citi Bike - New York & in- and out-flows & 01/07/2017 - 30/09/2017 & 1 hour & 80/10/10 \% & Xia et al., 2021 \cite{xia20213dgcn} \\
Seattle loop data - Seattle & traffic speeds & 01/11/2015 - 31/12/2015 & 5 minutes & 56/-/5 days & Yang et al., 2021 \cite{yang2021real} \\
\end{tabular}
\end{center}
\label{table:datasets}
\end{table*}%

\begin{figure*}[t]
    \centering
    \includegraphics[width=0.9\textwidth,trim={4cm 0.4cm 7cm 0.4cm},clip]{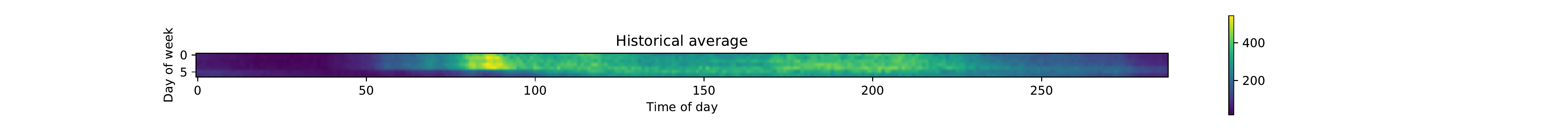}
    \caption{Weekly recurrent pattern for the PEMSD4 dataset. Day of week is encoded from 0 (Monday) to 6 (Sunday). Time of day is in 5min intervals. }
    \label{fig:stg-ncde-HA}
\end{figure*}

\begin{figure*}[t]
\vspace{-0.3cm}
    \hspace{0.7cm}
    \subfloat{%
      \includegraphics[width=0.39\textwidth,trim={1cm 0.4cm 3cm 0.4cm},clip]{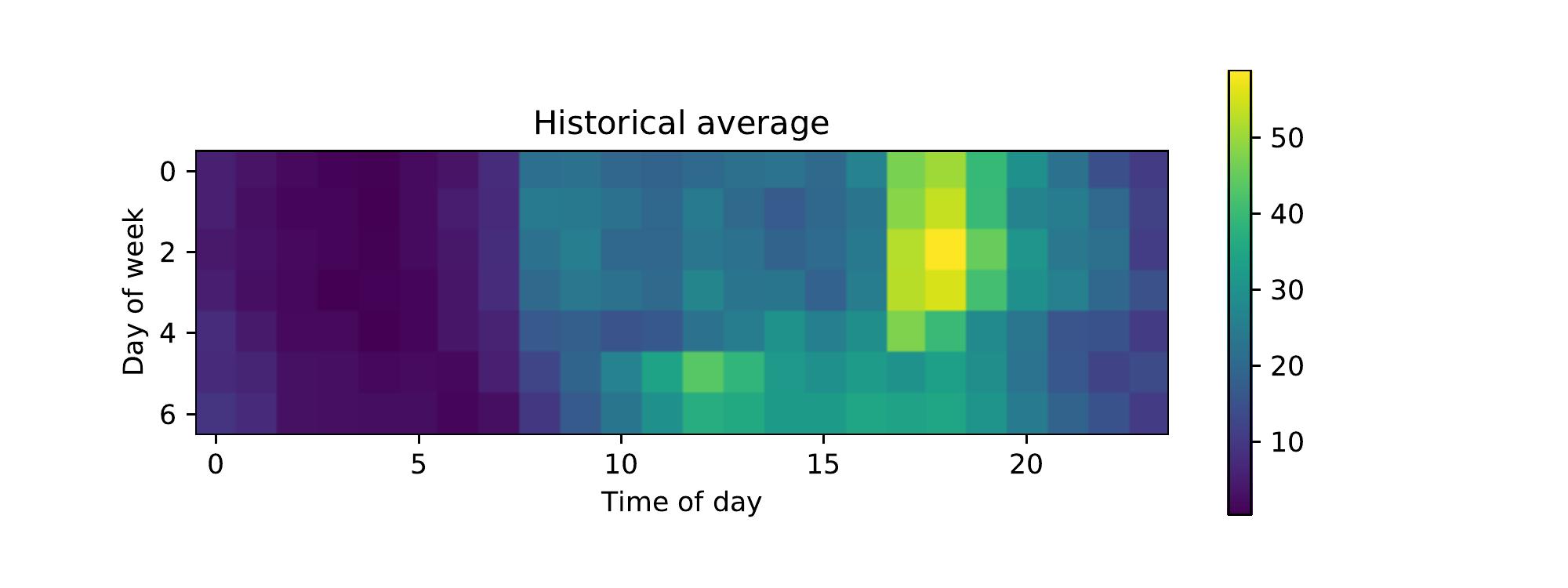}
    }
    \hspace{1cm}
    \subfloat{%
      \includegraphics[width=0.39\textwidth,trim={1cm 0.4cm 3cm 0.4cm},clip]{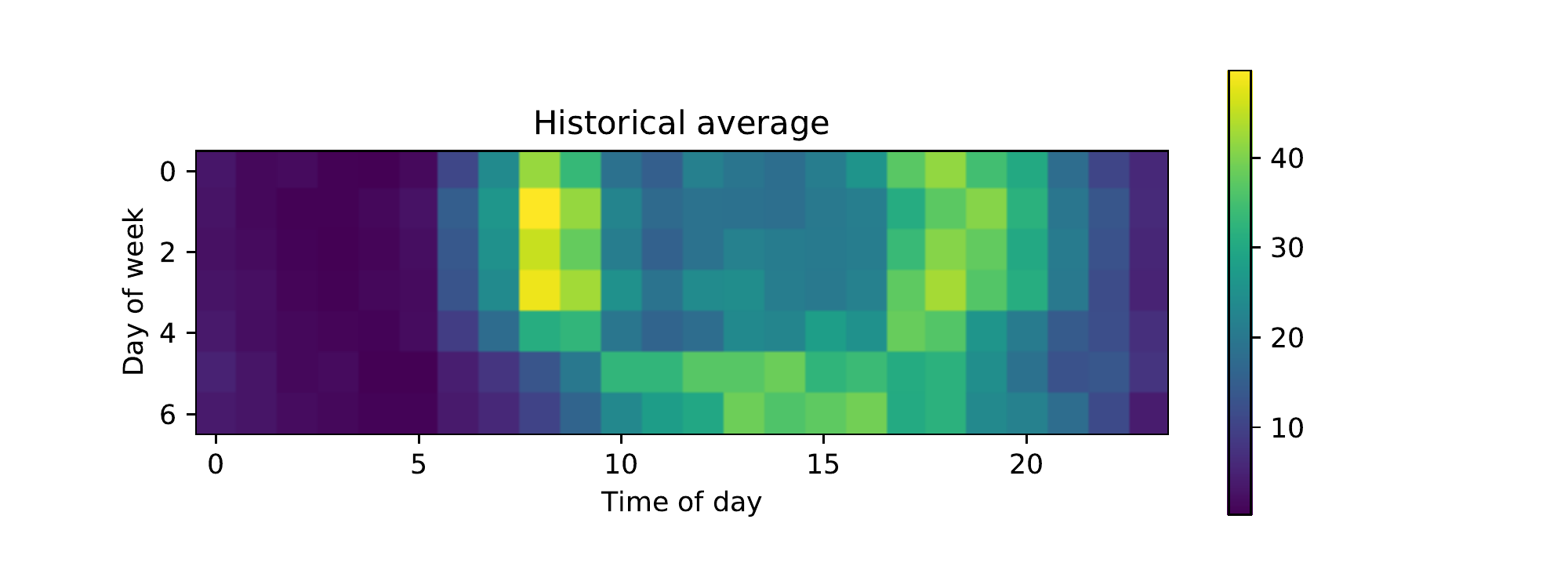}
    }
    \caption{Weekly recurrent pattern for the NYC CitiBike pickups (left) and dropoffs (right). Time of day is in 1h intervals. Monday=0 and Sunday=6.}
    \label{fig:3d-dgcn-HA}
\end{figure*}

We began by performing a search for recent publications on spatio-temporal transport prediction problems on top-tier journals and conferences that also make their source code and datasets publicly available online. This search yielded the 9 datasets (from 8 publications) that are summarized in Table~\ref{table:datasets}. After reproducing the results for the proposed approaches in each publication, we considered the exact same train/test set split and used the train set to compute the average weekly pattern for each location/area\footnote{When a validation set was also provided, it was concatenated to the train set and also used to compute the average weekly pattern.}, which is commonly referred in literature as the ``Historical Average'' (HA). Any holidays that were present in the data were treated as Sundays. Figures~\ref{fig:stg-ncde-HA} and \ref{fig:3d-dgcn-HA} show what this recurrent pattern looks like for the PEMSD4 and NYC CitiBike datasets, respectively, on two arbitrary locations/areas. The figures show clear patterns: Figure~\ref{fig:stg-ncde-HA} shows that during weekdays (0-4) traffic volumes are higher during the morning peak and lower during the night, while Figure~\ref{fig:3d-dgcn-HA} shows an area with higher in-flows on weekdays during the evening and higher out-flows on weekdays during the morning. 

Letting $a_t$ denote the value of the average recurrent pattern associated with time-step $t$, the original observations $y_t$ were transformed as follows: $y'_t = y_t - a_t$. The residuals $y'_t$ where then used to fit a linear regression model that regresses the value of $y'_t$ on the previous $h$ observations $\{y'_{t-1},\dots,y'_{t-h}\}$ based on the train set (effectively making this an autoregressive model of order $h$). Once fitted, this linear regression model can be applied to the test set to obtain predictions $\hat{y}'_t$, which can then be trivially converted back to the original domain by adding back the corresponding average recurrent pattern associated with each timestamp $t$: $\hat{y}_t = \hat{y}'_t + a_t$. We also considered a version of this approach where the standard deviation $s_t$ of the recurrent pattern was also included, thus resulting in $y'_t = (y_t - a_t) / s_t$. Although this can be relevant in some circumstances, we didn't observe significant improvements in forecasting error in the datasets considered. Once the final predictions $\hat{y}_t$ were obtained, we computed the same error metrics as provided in the corresponding publications such that the results are directly comparable. 

\section{Results}
\label{sec:results}

The source code to reproduce all the experiments in this paper is provided at \url{https://github.com/fmpr/mobility-baselines}. Tables~\ref{table:pemsd7m} to \ref{table:seattle} show the obtained results, where * is used to indicate results that were taken directly from the respective paper without reproducing them. ``HA" is used to denote the historical average estimated on the train set and applied to the test set, while ``HA+LR" denotes the simple baseline consisting of a linear regression on the residuals of the historical averages as described in Section~\ref{sec:methods}. Common baselines across the different tables include support vector regression (SVR), fully-connected neural networks (FNN), LSTM layers combined with a fully-connected output layer (FC-LSTM), vector autoregressive (VAR) models, ARIMA models, and gradient boosting (XGBoost). Other references to recent state-of-the-art methods are explained below. 

\textbf{PeMSD7(M).} Table~\ref{table:pemsd7m} shows the results of HA and HA+LR in comparison with the results in \cite{yu2017spatio} for traffic speed forecasting with forecasting horizons of 15, 30 and 45min using data from California. As it can be observed, a naive linear regression estimated on the previous $h=12$ residuals from the historical average (HA+LR) is able to obtain a lower RMSE for the 30 and 45min-ahead forecasting horizons, while being very competitive for the 15min forecasting horizon. 

\begin{table}[htp]
\caption{Results for PeMSD7(M) - traffic speeds in California. The * indicates results taken directly from \cite{yu2017spatio}.}
\begin{center}
\begin{tabular}{c|c|c|c}
& MAE & MAPE & RMSE\\
Model & 15/ 30/ 45 min & 15/ 30/ 45 min & 15/ 30/ 45 min\\
\hline
LSVR* & 2.50/ 3.63/ 4.54 & 5.81/ 8.88/ 11.50 & 4.55/ 6.67/ 8.28 \\
FNN* & 2.74/ 4.02/ 5.04 & 6.38/ 9.72/ 12.38 & 4.75/ 6.98/ 8.58 \\
FC-LSTM* & 3.57/ 3.94/ 4.16 & 8.60/ 9.55/ 10.10 & 6.20/ 7.03/ 7.51 \\
GCGRU* & 2.37/ 3.31/ 4.01 & 5.54/ 8.06/ 9.99 & 4.21/ 5.96/ 7.13 \\
STGCN(Cheb) & \textbf{2.25}/ \textbf{3.03}/ 3.57 & 5.26/ \textbf{7.33}/ 8.69 & \textbf{4.04}/ 5.70/ 6.77 \\
STGCN(1st) & 2.26/ 3.09/ 3.79 & \textbf{5.24}/ 7.39/ 9.12 & 4.07/ 5.77/ 7.03 \\
\hline
HA & 3.90 & 10.14 & 7.09 \\
HA+LR & 2.48/ 3.13/ \textbf{3.45} & 5.81/ 7.65/ \textbf{8.57} & 4.22/ \textbf{5.50}/ \textbf{6.10} \\
\end{tabular}
\end{center}
\label{table:pemsd7m}
\end{table}%

\textbf{Urban1.} Table~\ref{table:urban1} shows the results of HA and HA+LR in comparison with the results in \cite{lee2022ddp} for traffic speed forecasting with forecasting horizons of 30, 45 and 60min using data from South Korea. Although, in general, DDP-GCN provides the best results, the simple HA+LR baseline is extremely competitive. In fact, it even outperforms very popular approaches based on complex graph neural networks from the state of the art, such as DCRNN \cite{li2018diffusion} and STGCN \cite{yu2017spatio}. 

\begin{table}[h!]
\caption{Results for Urban1 - traffic speeds in South Korea. The * indicates results taken directly from \cite{lee2022ddp}. DCRNN and STGCN denote the approaches proposed in \cite{li2018diffusion} and \cite{yu2017spatio}, respectively.}
\begin{center}
\begin{tabular}{c|c|c|c}
& MAE & MAPE & RMSE\\
Model & 30/ 45/ 60 min & 30/ 45/ 60 min & 30/ 45/ 60 min\\
\hline
VAR* & 5.06/ 4.99/ 4.97 & 23.10/ 22.82/ 22.73 & 7.04/ 6.92/ 6.88  \\
LSVR* & 3.82/ 3.89/ 3.93 & 15.35/ 17.99/ 17.39 & 5.64/ 5.74/ 5.84 \\
ARIMA* & 3.49/ 3.79/ 4.04 & 15.40/ 16.85/ 18.09 & 5.28/ 5.65/ 5.94 \\
FC-LSTM* & 3.91/ 3.92/ 3.92 & 17.29/ 17.32/ 17.31 & 6.38/ 6.39/ 6.39 \\
DCRNN* & 3.17/ 3.46/ 3.73 & 13.52/ 14.83/ 15.95 & 4.94/ 5.30/ 5.61 \\
STGCN* & 3.07/ 3.42/ 3.80 & 14.38/ 16.72/ 19.37 & 4.57/ 4.83/ 5.04 \\
DDP-GCN & \textbf{3.00}/ \textbf{3.00}/ \textbf{2.99} & 13.57/ \textbf{13.56}/ \textbf{13.51} & \textbf{4.45}/ \textbf{4.45}/ \textbf{4.47} \\
\hline
HA & 3.18 & 14.19 & 4.79 \\
HA+LR & 3.04/ 3.10/ 3.13 & \textbf{13.39}/ 13.73/ 13.87 & 4.60/ 4.67/ 4.71 \\
\end{tabular}
\end{center}
\label{table:urban1}
\end{table}%

\textbf{NYC Citi Bike pickups and dropoffs.} Table~\ref{table:nyccitibike} shows the results of HA and HA+LR in comparison with the results in \cite{ye2020coupled} for forecasting bike pickups and dropoffs in NYC. The results correspond to a sequence-to-sequence approach, where the 12 previous observations are used to produce a forecast for the next 12 time steps. The results in Table~\ref{table:nyccitibike} then correspond to the average over the 12 forecasting horizons and consider both pickups and dropoffs. Interestingly, for this dataset, the HA+LR baseline outperforms the approach proposed in the original paper \cite{ye2020coupled}, as well as many other popular methods from the state of the art such as DCRNN \cite{li2018diffusion}, STGCN \cite{yu2017spatio}, STG2Seq \cite{bai2019stg2seq} and Graph WaveNet \cite{wu2019graph}. 

\begin{table}[h!]
\caption{Results for NYC Citi Bike - pickups and dropoffs in NYC. The * indicates results taken directly from \cite{ye2020coupled}.}
\begin{center}
\begin{tabular}{c|c|c}
Model & MAE & RMSE\\
\hline
XGBoost* & 2.469 & 4.050 \\
FC-LSTM* & 2.303 & 3.814 \\
DCRNN* & 1.895 & 3.209 \\
STGCN* & 2.761 & 3.604 \\
STG2Seq* & 2.498 & 3.984 \\
Graph WaveNet* & 1.991 & 3.294 \\
CCRNN & 1.740 & 2.838 \\
\hline
HA & \textbf{1.726} & 2.871\\
HA+LR & 1.738 & \textbf{2.758}\\
\end{tabular}
\end{center}
\label{table:nyccitibike}
\end{table}%

\textbf{PeMSD4.} Table~\ref{table:pemsd4} shows the results of HA and HA+LR in comparison with the results in \cite{choi2021graph} for forecasting traffic volumes using data from California. The results also correspond to a sequence-to-sequence approach, where the 12 previous observations are used to produce a forecast for the next 12 time steps. Table~\ref{table:pemsd4} shows the average results over the 12 forecasting horizons. Although the naive HA+LR baseline does not outperform the STG-NCDE approach proposed in \cite{choi2021graph}, it is very competitive and it again outperforms a substantial amount of other methods from the state of the art according to the results reported in \cite{choi2021graph}. The reader is referred to the original paper for a detailed description of all the methods considered in Table~\ref{table:pemsd4}. 

\begin{table}[h!]
\caption{Results for PeMSD4 - traffic volumes in California. The * indicates results taken directly from \cite{choi2021graph}.}
\begin{center}
\begin{tabular}{c|c|c|c}
Model & MAE & MAPE & RMSE\\
\hline
ARIMA* & 33.73 & 24.18 & 48.80 \\
VAR* & 24.54 & 17.24 & 38.61 \\
FC-LSTM* & 26.77 & 18.23 & 40.65 \\
TCN* & 23.22 & 15.59 & 37.26 \\
GRU-ED* & 23.68 & 16.44 & 39.27 \\
DSANet* & 22.79 & 16.03 & 35.77 \\
STGCN* & 21.16 & 13.83 & 34.89 \\
DCRNN* & 21.22 & 14.17 & 33.44 \\
GraphWaveNet* & 24.89 & 17.29 & 39.66 \\
ASTGCN(r)* & 22.93 & 16.56 & 35.22 \\
MSTGCN* & 23.96 & 14.33 & 37.21 \\
STG2Seq* & 25.20 & 18.77 & 38.48 \\
LSGCN* & 21.53 & 13.18 & 33.86 \\
STSGCN* & 21.19 & 13.90 & 33.65 \\
AGCRN* & 19.83 & 12.97 & 32.26 \\
STFGNN* & 20.48 & 16.77 & 32.51 \\
STGODE* & 20.84 & 13.77 & 32.82 \\
Z-GCNETs* & 19.50 & 12.78 & 31.61 \\
STG-NCDE & \textbf{19.21} & \textbf{12.76} & \textbf{31.09} \\
\hline
HA & 26.26 & 17.07 & 42.87\\
HA+LR & 20.03 & 13.39 & 32.73\\
\end{tabular}
\end{center}
\label{table:pemsd4}
\end{table}%

\textbf{SZ-taxi.} Table~\ref{table:sztaxi} shows the results of HA and HA+LR in comparison with the results in \cite{zhao2019t} for traffic speed forecasting with forecasting horizons of 15, 30, 45 and 60min using data from Shenzhen.   In this particular case, re-running the code provided by the authors yielded different results than the ones reported in \cite{zhao2019t}. Our comparison is then based on the code and data provided, and not on the results from the paper. As Table~\ref{table:sztaxi} shows, the simple HA+LR baseline outperforms the T-GCN approach proposed in \cite{zhao2019t} for all forecasting horizons considered. 

\begin{table}[h!]
\caption{Results for SZ-taxi - traffic speeds in Shenzhen from taxi trajectories. The * indicates results taken directly from \cite{zhao2019t}.}
\begin{center}
\begin{tabular}{c|c|c}
& MAE & RMSE\\
Model & 15/ 30/ 45/ 60 min & 15/ 30/ 45/ 60 min\\
\hline
T-GCN & 4.517/ 4.572/ 4.621/ 4.671 & 5.997/ 6.034/ 6.064/ 6.100 \\
\hline
HA & 4.630 & 6.463 \\
HA+LR & \textbf{3.464}/ \textbf{3.507}/ \textbf{3.534}/ \textbf{3.554} & \textbf{4.998}/ \textbf{5.057}/ \textbf{5.091}/ \textbf{5.115} \\
\end{tabular}
\end{center}
\label{table:sztaxi}
\end{table}%

\textbf{METR-LA.} Table~\ref{table:metrla} shows the results of HA and HA+LR in comparison with the results in \cite{li2018diffusion} for traffic speed forecasting with forecasting horizons of 15, 30 and 60min using data from LA. Although DCRNN obtains the best results for most forecasting horizons, HA+LR is shown to be the most competitive method of all the other methods considered. 

\begin{table}[htp]
\caption{Results for METR-LA - traffic speeds in LA. The * indicates results taken directly from \cite{li2018diffusion}.}
\begin{center}
\begin{tabular}{c|c|c|c}
& MAE & MAPE & RMSE\\
Model & 15/ 30/ 60 min & 15/ 30/ 60 min & 15/ 30/ 60 min\\
\hline
ARIMA* & 3.99/ 5.15/ 6.90 & 9.6/ 12.7/ 17.4 & 8.21/ 10.45/ 13.23 \\
VAR* & 4.42/ 5.41/ 6.52 & 10.2/ 12.7/ 15.8 & 7.89/ 9.13/ 10.11 \\
SVR* & 3.99/ 5.05/ 6.72 & 9.3/ 12.1/ 16.7 & 8.45/ 10.87/ 13.76 \\
FNN* & 3.99/ 4.23/ 4.49 &  9.9/ 12.9/ 14.0 & 7.94/ 8.17/ 8.69 \\
FC-LSTM* & 3.44/ 3.77/ 4.37 & 9.6/ 10.9/ 13.2 & 6.30/ 7.23/ 8.69 \\
DCRNN & \textbf{2.77}/ \textbf{3.15}/ \textbf{3.60} & \textbf{7.3}/ \textbf{8.8}/ \textbf{10.5} & \textbf{5.38}/ \textbf{6.45}/ 7.59 \\
\hline
HA & 4.19 & 13.0 & 7.84 \\
HA+LR & 3.28/ 3.68/ 4.02 & 8.8/ 10.4/ 11.9 & 5.71/ 6.60/ \textbf{7.32} \\
\end{tabular}
\end{center}
\label{table:metrla}
\end{table}%

\textbf{PEMS-BAY.} Table~\ref{table:pemsbay} shows the results of HA and HA+LR in comparison with the results in \cite{li2018diffusion} for traffic speed forecasting with forecasting horizons of 15, 30 and 60min using another dataset from California. Interestingly, in this case, the simple HA+LR baseline is shown to obtain the best results in terms or RMSE for all forecasting horizons. 

\begin{table}[htp]
\caption{Results for PEMS-BAY - traffic speeds in California. The * indicates results taken directly from \cite{li2018diffusion}.}
\begin{center}
\begin{tabular}{c|c|c|c}
& MAE & MAPE & RMSE\\
Model & 15/ 30/ 60 min & 15/ 30/ 60 min & 15/ 30/ 60 min\\
\hline
ARIMA* & 1.62/ 2.33/ 3.38 & 3.5/ 5.4/ 8.3 & 3.30/ 4.76/ 6.50 \\
VAR* & 1.74/ 2.32/ 2.93 & 3.6/ 5.0/ 6.5 & 3.16/ 4.25/ 5.44 \\
SVR* & 1.85/ 2.48/ 3.28 & 3.8/ 5.5/ 8.0 & 3.59/ 5.18/ 7.08 \\
FNN* & 2.20/ 2.30/ 2.46 & 5.2/ 5.4/ 5.9 & 4.42/ 4.63/ 4.98 \\
FC-LSTM* & 2.05/ 2.20/ 2.37 & 4.8/ 5.2/ 5.7 & 4.19/ 4.55/ 4.96 \\
DCRNN & \textbf{1.38}/ \textbf{1.74}/ \textbf{2.07} & \textbf{2.9}/ \textbf{3.9}/ \textbf{4.9} & 2.95/ 3.97/ 4.74 \\
\hline
HA & 2.58 & 6.1 & 5.04 \\
HA+LR & 1.54/ 1.91/ 2.22 & 3.2/ 4.3/ 5.1 & \textbf{2.93/ 3.83/ 4.45} \\
\end{tabular}
\end{center}
\label{table:pemsbay}
\end{table}%

\textbf{NYC Bike in- and out-flows.} Table~\ref{table:inoutflows} shows the results of HA and HA+LR in comparison with the results in \cite{xia20213dgcn} for predicting in- and out-flows in NYC with forecasting horizons of 1, 2 and 3 hours. As it turns out, although for an 1-hour horizon 3DGCN tends to yield better results, for the longer horizons the simple HA+LR baseline is able to outperform it. Moreover, HA+LR is once again shown to outperform popular approaches from the state of the art such as DCRNN \cite{li2018diffusion} and STGCN \cite{yu2017spatio} for this dataset according to the results provided by the authors of \cite{xia20213dgcn}. 

\begin{table}[htp]
\caption{Results for NYC Bike in- and out-flows. The * indicates results taken directly from \cite{xia20213dgcn}.}
\begin{center}
\begin{tabular}{c|c|c}
& MAE & RMSE\\
Model & 1h/ 2h/ 3h & 1h/ 2h/ 3h\\
\hline
ARIMA* & 10.41/ 11.84/ 13.00 & 19.14/ 21.76/ 23.90 \\
STGCN* & 6.49/ 7.06/ 7.94 & 11.73/ 12.93/ 15.37 \\
DCRNN* & 5.88/ 6.19/ 7.72 & 9.85/ 10.39/ 12.37 \\
STGNN* & 5.79/ 6.00/ 7.56 & 9.80/ 9.98/ 11.91 \\
MVGCN* & 5.65/ 7.72/ 8.00 & 9.64/ 13.53/ 13.93 \\
3DGCN & \textbf{4.81}/ 5.61/ 6.99 & \textbf{7.76}/ \textbf{9.49}/ 11.74 \\
\hline
HA &  5.97 & 11.04 \\
HA+LR & 5.10/ \textbf{5.45}/ \textbf{5.56} & 8.72/ 9.69/ \textbf{10.04}\\
\end{tabular}
\end{center}
\label{table:inoutflows}
\end{table}%

\textbf{Seattle traffic speeds.} Table~\ref{table:seattle} shows the results of HA and HA+LR in comparison with the results in \cite{yang2021real} for predicting traffic speeds in Seattle with a forecasting horizon of 5 minutes. Although the main focus of \cite{yang2021real} is on handling missing data and how different missing rates affect prediction performance, the authors also provide results for the different methods studied under no missing data. We compare HA and HA+LR with those results. As Table~\ref{table:seattle} shows, the simple HA+LR baseline is able to significantly outperform all the other approaches considered in \cite{yang2021real}. 

\begin{table}[htp]
\caption{Results for Seattle traffic speed data. The * indicates results taken directly from \cite{yang2021real}.}
\begin{center}
\begin{tabular}{c|c|c}
& MAPE & RMSE\\
\hline
BTMF* & 7.70 & 4.59\\
TRMF-GRMF* & 8.20 & 4.83\\
TRMF-ALS* & 8.36 & 4.96\\
Linear-LSTM-ReMF* & 8.01 & 4.59\\
BiLSTM-GL-ReMF & 7.83 & 4.50\\
LSTM-ReMF & 7.64 & 4.42\\
LSTM-GL-ReMF & 7.64 & 4.43\\
\hline
HA & 12.5 & 9.82 \\
HA+LR & \textbf{5.5} & \textbf{3.95}\\
\end{tabular}
\end{center}
\label{table:seattle}
\end{table}%

\section{Discussion}

\textbf{Stationarity and recurrent weekly patterns.} The results in Section~\ref{sec:results} make the importance of accounting for the recurrent weekly patterns very clear. Although computing the residuals to the recurrent weekly pattern (the so-called ``historical average") does not guarantee stationarity of the resulting time series, it is certainly a step in the right direction. For datasets that also include a trend component (e.g., when forecasting demand for a new mobility service whose user base is growing over time), one may also wish to consider detrending approaches \cite{hyndman2018forecasting}, or use seasonal differencing instead. 

\textbf{Baselines.} The results in Section~\ref{sec:results} also highlight the need for strong baselines that ideally are also simple to implement. When using only other state-of-the-art approaches as a reference, it is crucial that enough time is spent on carefully tuning them so that the comparison is fair, which can be difficult to achieve in practice. For example, when reproducing the results from other papers, we noticed that some ``ARIMA" baselines actually corresponded to ARIMA(1,0,0) models (i.e., effectively an autoregressive model of order 1), and that some historical average (``HA") baselines reported in the papers actually corresponded to the average over the last $h$ observations. On the other hand, as empirically demonstrated, the HA+LR baseline is trivial to implement, takes seconds to compute, requires no hyper-parameter tuning, and can often achieve comparable results to many complex state-of-the-art approaches. Furthermore, it can be extended to account for spatial correlations by considering a vector autoregressive approach instead (or even a graph neural network), or to account for non-linear relationships by introducing a more flexible machine learning model in the place of linear regression. 

\textbf{Benchmarks.} Most of the issues that this short paper highlights are aggravated by the inexistence of standard reference benchmarks. Our analysis of the literature on spatio-temporal forecasting in transportation revealed that most papers tend to use their own dataset(s) with their own experimental setup (train/val/test split, forecasting horizons, evaluation metrics, etc.), which, more often than not, are not made publicly available. This makes direct comparisons across different approaches more difficult to achieve than necessary, since it requires the authors to re-implement or, in the best-case scenario, perform hyper-parameter tuning to the new dataset, which significantly increases the risk of implementation mistakes or the hyper-parameter tuning not being sufficiently thorough. 

Moreover, different datasets have different properties. The ``no free lunch" theorem \cite{wolpert2002supervised} tells us that all machine learning algorithms are equally effective when averaged across all possible prediction problems. It is therefore important that multiple datasets are considered in order to empirically demonstrate the advantages of a newly proposed method. To take a step in this direction, we provide a centralized repository\footnote{\url{https://github.com/fmpr/mobility-baselines}} with code for preparing the experimental setups (train/val/test split, forecasting horizons, evaluation metrics, etc.) for the 9 publicly-available datasets considered in this paper, so that future works can easily compare their approaches with the results from Tables~\ref{table:pemsd7m} to \ref{table:seattle}. 

\textbf{Sharing code and datasets.} Together with the use of open benchmarks, other good practices such as the sharing of code and datasets has massively accelerated the pace of progress in other fields such as computer vision and natural language processing. It is unfortunate that the transportation community has not yet widely embraced these good practices, since they can facilitate the comparison of approaches and speed up scientific progress. 

\textbf{Relevancy.} Are these improvements relevant in practice? Is the field making progress, or are we stuck in a loop of small incremental contributions and repeating the same mistakes? Despite the abundance of research contributions on spatio-temporal forecasting for transportation in recent years, there is a serious risk that the community is loosing track of the practical problems that these methods are trying to solve. Does an improvement of 0.5 in MAE when forecasting traffic speeds have a relevant impact on traffic management in order to justify a complex model that requires several hours to train and multiple days to tune its hyper-parameters? Is the $\mbox{CO}_2$ fingerprint that we are causing due to all this computation perhaps larger than the savings in congestion and other efficiency gains? 

\textbf{Evaluation metrics.} The relevancy problem tends to be aggravated by the focus on error metrics like MAE and RMSE. Although it can be difficult to achieve, the community should try to progressively move away from these error metrics towards other KPIs that are better proxies of societal impact, such as the number of hours spent in traffic, $\mbox{CO}_2$ emissions, quality of service, etc. When the predictions of a model are used as inputs for another downstream task, the empirical evaluation should include the improvements on that task. But even when solely error metrics like MAE and RMSE are used, we argue that the emphasis should be on abnormal conditions, such as the ones caused by incidents, special events, extreme weather, etc., since these are the periods that stress the transportation system the most and, therefore, are most likely to have a big impact on the users. As the results in this paper highlight, forecasting traffic speeds or transport demand with reasonable accuracy does not require extremely complex models. However, spatio-temporal forecasting under abnormal conditions can be extremely challenging.  

\textbf{Spatial correlations.} Although the results in this paper highlight the importance of temporal correlations and average weekly patterns, the spatial correlations should not be neglected. In fact, the latter can have a significant impact on forecasting error, especially when shorter forecasting horizons. This was also observed during our empirical studies and can be verified in the results provided in Section~\ref{sec:results}. As longer forecasting horizons are considered, the impact of modeling spatial correlations on forecasting error gradually decreases. 

\bibliographystyle{IEEEtran}
\bibliography{bibliography}

%
%
%
%

\vfill

\end{document}